\documentclass[journal]{IEEEtran}
\usepackage{subcaption}
\usepackage{amsmath,amsfonts}
\usepackage{algorithmic}
\usepackage{array}
\usepackage{cite}
\usepackage{textcomp}
\usepackage{stfloats}
\usepackage{url}
\usepackage{verbatim}
\usepackage{graphicx}
\usepackage{xcolor}
\usepackage{hyperref}
\usepackage{soul}
\usepackage{verbatim}
\usepackage[table]{xcolor}
\usepackage{multirow}
\usepackage{booktabs}

\usepackage{orcidlink}
\usepackage{pifont}

\ifCLASSINFOpdf

\else

\fi

\begin{document}

\title{Dynamic Span Interaction and Graph-Aware Memory for Entity-Level Sentiment Classification}

\author{
Md. Mithun Hossain \orcidlink{0009-0001-4883-1802}, 
Sanjara \orcidlink{0009-0002-6315-4683}, 
Md.~Shakil~Hossain \orcidlink{0009-0009-1584-3282}, and
Sudipto~Chaki \orcidlink{0000-0002-7286-6722}

\thanks{Md. Mithun Hossain, Sanjara, Md. Shakil Hossain, and Sudipto Chaki are with the Department of Computer Science and Engineering, Bangladesh University of Business and Technology, Dhaka 1216, Bangladesh; e-mail: (mhosen751@gmail.com, sanjaratunola@gmail.com, shakilhosen3.1416@gmail.com, sudiptochakibd@gmail.com).}

\thanks{Corresponding Author: Md. Mithun Hossain (e-mail: mhosen751@gmail.com)}

\thanks{Manuscript received Month xx, 20xx; revised Month xx, 20xx.}}

\markboth{Journal of \LaTeX\ Class Files,~Vol.~14, No.~8, August~2015}%
{M. Mithun Hossain \MakeLowercase{\textit{et al.}}: Dynamic Span Interaction and Graph-Aware Memory for Entity-Level Sentiment Classification}

\maketitle

\begin{abstract}
Entity-level sentiment classification involves identifying the sentiment polarity linked to specific entities within text. This task poses several challenges: effectively modeling the subtle and complex interactions between entities and their surrounding sentiment expressions; capturing dependencies that may span across sentences; and ensuring consistent sentiment predictions for multiple mentions of the same entity through coreference resolution. Additionally, linguistic phenomena such as negation, ambiguity, and overlapping opinions further complicate the analysis. These complexities make entity-level sentiment classification a difficult problem, especially in real-world, noisy textual data. To address these issues, we propose SpanEIT, a novel framework integrating dynamic span interaction and graph-aware memory mechanisms for enhanced entity-sentiment relational modeling. SpanEIT builds span-based representations for entities and candidate sentiment phrases, employs bidirectional attention for fine-grained interactions, and uses a graph attention network to capture syntactic and co-occurrence relations. A coreference-aware memory module ensures entity-level consistency across documents. Experiments on FSAD, BARU, and IMDB datasets show SpanEIT outperforms state-of-the-art transformer and hybrid baselines in accuracy and F1 scores. Ablation and interpretability analyses validate the effectiveness of our approach, underscoring its potential for fine-grained sentiment analysis in applications like social media monitoring and customer feedback analysis.
\end{abstract}


\IEEEpeerreviewmaketitle

\section{Introduction}  
\label{sec1}
Sentiment analysis is a fundamental task in natural language processing (NLP) that focuses on extracting opinions from large-scale textual data, offering valuable insights into public attitudes and trends~\cite{liu2022sentiment,cambria2018sentiment}. Conventional methods usually give whole documents or sentences a coarse polarity label, such as positive, neutral, or negative~\cite{pang2008opinion, socher2013recursive}. However, these methods often overlook the nuanced details present when multiple entities or conflicting sentiments coexist within the same text. To refine this coarse-grained analysis, Aspect-Based Sentiment Analysis (ABSA) has emerged as a more fine-grained paradigm that targets the identification of sentiment polarity toward specific textual elements or aspects~\cite{pontiki2016semeval}. ABSA has gained a lot of traction in areas like social media monitoring, hospitality, and product reviews where stakeholders need detailed sentiment insights at the level of service features or product attributes. For instance, "screen quality" or "battery life" in smartphone reviews are examples~\cite{ma2017interactive, wang2018target, tang2016aspect}. Recent advances in ABSA leverage deep learning architectures such as recurrent neural networks~\cite{wang2016attention}, convolutional neural networks~\cite{tang2015effective}, and transformer-based models~\cite{devlin2019bert}, often enhanced with attention mechanisms that dynamically focus on sentiment-bearing contextual cues relevant to each aspect~\cite{ma2017interactive, ruidan2020exploiting}. Despite these advancements, most ABSA models rely on predefined aspect categories and typically address only one aspect per sentiment expression, which limits their flexibility and applicability in real-world texts where aspect boundaries can be ambiguous or multiple aspects overlap. Furthermore, ABSA approaches generally concentrate on intra-sentence analysis and may struggle with complex linguistic phenomena such as co-reference, negation, and sentiment interactions that span multiple sentences. These challenges constrain their effectiveness in practical scenarios involving multiple interacting entities and overlapping opinions~\cite{liu2015fine, hossain2025crosgrpsabs}.

ABSA's capabilities are limited when dealing with complicated real-world texts, where attitudes are frequently conveyed toward various things that may overlap or co-refer across phrases, even with its advancements.  For more actionable insights in many applications, including customer relationship management, social media opinion mining, and targeted marketing, it is essential to comprehend sentiment that is particularly directed toward individual entities (people, organizations, or goods) rather than general features~\cite{li2017learning, Cambria2020senticnet}. Moreover, the presence of coreference, where the same entity is mentioned multiple times under different surface forms, poses challenges for consistent sentiment attribution that most ABSA frameworks do not adequately address~\cite{iqbal2025coreference, chen2024document, MITHUNHOSSAIN}. Additionally, the syntactic and semantic relationships among entities and sentiment expressions often span beyond sentence boundaries and involve long-distance dependencies that standard sequential models struggle to capture~\cite{jin2024span, huang2021entity}. These challenges motivate the need for a more comprehensive entity-aware sentiment analysis framework that can dynamically model fine-grained interactions between entity mentions and their sentiment contexts, incorporate structural linguistic information, and maintain entity-level consistency throughout the document.

Motivated by these challenges, we propose \textbf{SpanEIT}, a novel span-level entity-sentiment interaction transformer framework specifically designed to tackle these complexities. SpanEIT constructs rich semantic representations for candidate entity and sentiment spans and applies a dynamic bidirectional attention mechanism to capture fine-grained entity-sentiment interactions. To further enhance relational reasoning, it incorporates a span-level graph encoding syntactic and co-occurrence relationships, processed using graph attention networks (GATs). Moreover, a co-reference-aware memory module maintains and updates entity-specific contextual embeddings across multiple mentions, ensuring coherent sentiment attribution throughout the document. This combination of dynamic span interaction, graph-based reasoning, and memory-augmented consistency makes SpanEIT uniquely capable of delivering accurate and context-sensitive entity-level sentiment classification.

The principal contributions of this work are as follows:

\begin{itemize}
    \item We propose a novel span-based interaction framework that models detailed, bidirectional relationships between entity and sentiment spans, enabling precise sentiment attribution at the entity level.
    \item We incorporate a span-aware graph attention mechanism that captures both syntactic and semantic relations between spans, improving the model's capacity for structured reasoning.
    \item We introduce a coreference-aware memory module that maintains coherent representations for recurring entities, ensuring contextual consistency in sentiment predictions across the document.
\end{itemize}

The remainder of this paper is structured as follows. Section~\ref{sec2} reviews related work on sentiment analysis, with a focus on entity-level and span-based methods. Section~\ref{sec3} details the SpanEIT architecture, including core modules for span interaction, graph-based reasoning, and memory-aware context tracking. Section~\ref{sec4} presents experimental settings, datasets, evaluation metrics, and results. Section~\ref{sec5} discusses the limitations of our approach and offers directions for future research. Finally, Section~\ref{sec6} concludes the paper by summarizing key findings and contributions.

\section{Literature Review}
\label{sec2}

\textbf{Evolution of Entity-Level Sentiment Analysis:}
Sentiment analysis has seen significant advancements across sentence-level \cite{phan2024fuzzy} and document-level tasks \cite{wasi2024skeds}, primarily focusing on identifying the overall sentiment polarity of textual data \cite{tang2015joint}. Earlier approaches utilized bag-of-words models \cite{graff2025bag}, handcrafted features, and traditional classifiers such as SVM and Na\"ive Bayes \cite{vidyashree2024tweet}, but these methods lacked granularity and struggled with aspect-based or entity-targeted sentiment extraction \cite{jian2024retrieval}. The evolution toward Aspect-Based Sentiment Analysis (ABSA) addressed these limitations by associating sentiments with specific aspects \cite{ouyang2024aspect}, thus enabling the development of fine-grained entity-level sentiment classification (ESC) systems. ESC aims to determine sentiment polarities directed at individual entities within a sentence or document. Initial systems relied on pipeline architectures \cite{yu2019entity}, combining Named Entity Recognition with rule-based or statistical classifiers \cite{engonopoulos2011els}. With deep learning advancements, architectures such as RNNs \cite{yang2019multi}, LSTMs \cite{yu2019entity}, and Transformers \cite{song2019attention} improved representation learning. However, as noted by Zhao et al. \cite{zhao2024enhancing}, most of these models process text sequentially, which hinders their ability to model long-range interactions between sentiment expressions and entities. Furthermore, handling multiple entities with ambiguous sentiment remains a challenge.

\textbf{Span- and Graph-Based Modeling:}
Span-based approaches, which model contiguous text segments rather than individual tokens, have proven useful in diverse tasks including coreference resolution, event extraction, and question answering \cite{lv2021span}. Unlike token-level methods, span representations enable more interpretable reasoning \cite{tang2020span}. While such approaches have shown promise in detecting opinion targets \cite{zhang2024simple}, their integration into ESC remains underdeveloped. Recent studies, such as Xu et al. \cite{xu2025span}, suggest that span-pair interaction modeling can enhance contextual alignment, yet few models incorporate this for sentiment tasks \cite{du2024span}. Graph Neural Networks (GNNs) and Graph Attention Networks (GATs) have demonstrated strong potential in representing syntactic and semantic dependencies \cite{su2024enhanced, liang2022aspect}. By leveraging graph structures, these models can capture cross-sentence relationships and entity-aspect dependencies more effectively \cite{yin2024textgt}. For instance, Zhang et al. \cite{zhang2023enhanced} employed syntactic GCNs for ABSA, improving generalization. However, most graph-based sentiment models remain focused on token-level granularity and do not capture span-level semantics, a gap still open for exploration.

\textbf{Coreference and Memory Mechanisms for Contextual Consistency:}
Preserving sentiment coherence across recurring mentions of an entity requires both coreference resolution and memory mechanisms. Bohnet et al. \cite{bohnet2023coreference} demonstrate that seq2seq models can enhance entity consistency, yet Liu et al. \cite{liu2023brief} note that integration into sentiment classification pipelines remains limited. To address long-range dependencies, models like GRU-based memory networks \cite{wu2024xlnet} and commonsense-enhanced adapters \cite{lu2023commonsense} offer promising directions, but they are rarely applied to ESC, where maintaining and updating entity states is critical. Despite these advances, several research gaps persist: there is a lack of explicit modeling for bidirectional interactions between entity and sentiment spans, reducing the capacity for fine-grained sentiment comprehension; graph-based relational reasoning at the span level, which could integrate both syntactic and semantic cues, is underutilized; and limited incorporation of coreference and memory mechanisms hinders consistent sentiment attribution across co-referent entities. To address these challenges, the proposed SpanEIT framework integrates span-level interaction modeling, graph-based reasoning, and coreference-aware memory. It aligns entity and sentiment spans dynamically, enables structured reasoning over syntactic and semantic relations, and ensures consistent sentiment attribution across coreferent mentions.

\section{Proposed Methodology}
\label{sec3}
\subsection{Problem Formulation}

Entity-level sentiment analysis aims to identify the sentiment polarity expressed toward specific entities within a given sentence. Formally, let \( S = (w_1, w_2, \dots, w_n) \) denote a sentence consisting of \( n \) tokens, and let \( E = \{e_1, e_2, \dots, e_m\} \) represent the set of entities recognized in \( S \), where each entity \( e_i \) is characterized by its token span and entity type \( t_i \). The objective is to learn a mapping function

\begin{equation} \label{eq:mapping}
f: (S, e_i, t_i) \rightarrow y_i,
\end{equation}

where the sentiment label for entity \( e_i \) is indicated by \( y_i \in \{ \text{positive}, \text{neutral}, \text{negative} \} \). This calls for modeling the relationships between entities and sentiment-bearing spans, accurate representation of entity borders and kinds, and a thorough comprehension of the sentence context. Consistency in sentiment assignment also requires addressing issues like coreference resolution, which occurs when numerous mentions may relate to the same underlying item.

\begin{figure}[htbp]
  \centering
  \includegraphics[scale=0.56]{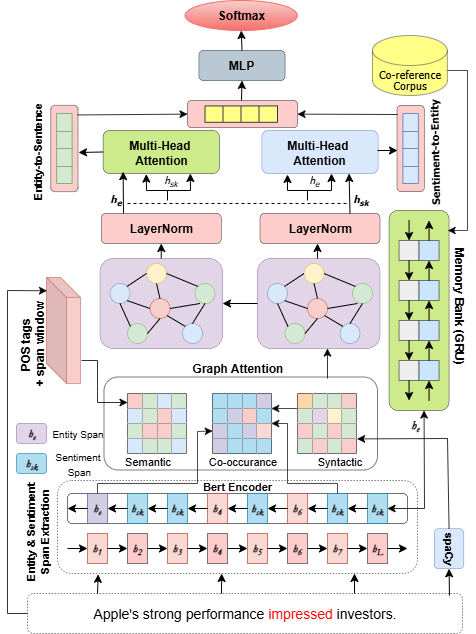}
    \caption{\small SpanEIT architecture for entity-level sentiment analysis, extracting entity and sentiment spans from text and building a combined syntactic-semantic graph processed by Graph Attention Networks to generate context-aware embeddings. Multi-head cross-attention captures interactions between entities and sentiments, while a memory bank maintains coreference-aware representations. The fused features are classified via a Multi-Layer Perceptron for fine-grained sentiment prediction. }
  \label{architecture}
\end{figure}

Given a labeled dataset

\begin{equation} \label{eq:dataset}
\mathcal{D} = \{(S^{(k)}, E^{(k)}, Y^{(k)})\}_{k=1}^N,
\end{equation}

where each \( S^{(k)} \) is a sentence, \( E^{(k)} \) denotes the set of entities in \( S^{(k)} \), and \( Y^{(k)} \) are the corresponding sentiment labels, the goal is to optimize the mapping \( f \) in Equation~\eqref{eq:mapping} to accurately predict \( y_i \) for all entities across the dataset in Equation~\eqref{eq:dataset}. This formulation motivates the need for advanced architectures that can leverage contextualized embeddings, span-level representations, graph-structured relationships, and memory components to capture the nuanced dependencies present in entity-level sentiment classification. An overview of our proposed SpanEIT model, designed to address these challenges, is illustrated in Figure~\ref{architecture}.

\subsection{Input Representation and Contextual Embedding}
To obtain rich, context-sensitive representations of input text, we leverage a pre-trained Bidirectional Encoder Representations from Transformers (BERT) model~\cite{devlin2019bert}. For each input sentence $S = (w_1, w_2, \dots, w_n)$, we tokenize and encode the text using BERT to produce contextualized embeddings $\mathbf{H} = (\mathbf{h}_1, \mathbf{h}_2, \dots, \mathbf{h}_n)$, where each $\mathbf{h}_i \in \mathbb{R}^d$ captures semantic and syntactic context for token $w_i$. These embeddings serve as the foundational features for all subsequent modules, including span extraction, graph construction, attention mechanisms, and memory components. By initializing with BERT, our architecture benefits from deep, language-aware features that enhance the modeling of entity-sentiment interactions in later stages.

\subsection{POS Tagging and Span Extraction}

Building on the BERT-derived embeddings, we tokenize each input sentence, \( S = (w_1, w_2, \dots, w_n) \), and apply part-of-speech (POS) tagging using a high-accuracy syntactic parser such as spaCy. The POS tags provide essential linguistic cues for identifying both entity mentions and sentiment-bearing expressions within the text. Leveraging these tags, as well as predefined contextual windows and linguistic heuristics, we systematically extract two categories of spans: (i) {entity spans} \( E = \{ e_1, e_2, \dots, e_m \} \), which typically correspond to noun phrases or named entities (e.g., ``Apple’s'', ``investors''), and (ii) {sentiment spans} \( S_p = \{ s_1, s_2, \dots, s_k \} \), which capture key adjectives, verbs, or phrases that convey evaluative meaning (e.g., ``strong'', ``performance''). Each span is defined as a contiguous subsequence of tokens, effectively localizing linguistically salient units that are central to sentiment reasoning.
This span extraction process transforms the flat token sequence \( S \) into two structured sets of spans, allowing our model to focus selectively on the most informative parts of the sentence. These extracted spans serve as the foundation for subsequent representation learning and enable explicit modeling of entity-sentiment interactions. By grounding the model’s reasoning in these linguistically meaningful units, we facilitate both improved sentiment attribution and more interpretable model behavior in downstream tasks.

\subsection{Graph Construction: Syntactic and Semantic}

From the extracted tokens and spans, we constructed two complementary graphs to comprehensively capture linguistic relationships. First, we built the syntactic graph 
$G_{syn} = (V, E_{syn})$ by encoding grammatical dependencies such as adjectival modifiers and possessives derived from the sentence’s dependency parse tree. This graph explicitly represents the structural relationships between words and spans, providing essential syntactic context. Next, to complement this, we constructed a semantic co-occurrence graph 
$G_{sem} = (V, E_{sem})$ that captures statistical co-occurrence strengths between entities and sentiment spans. This graph uncovers latent semantic associations that are not directly reflected in syntactic links. We then combined these two graphs into a unified graph 
$G = (V, E_{syn} \cup E_{sem})$. This unified representation, incorporating both syntactic and semantic edges, was fed into the graph attention network layers, enabling the model to jointly leverage explicit syntactic structures and implicit semantic proximities for richer, context-aware node embeddings.

\subsection{Graph Attention Network (GAT) Layers}

Our model then passed the combined graph $G$ through Graph Attention Network (GAT) layers \cite{velivckovic2017graph} to compute context-aware embeddings for each node. At each layer $l$, as defined in Equation~\eqref{eq:gat_update}, we updated the embedding $h_i^{(l)}$ of a node $v_i$ by aggregating information from its immediate neighbors $N(i)$. Each neighbor’s embedding $h_j^{(l)}$ was first transformed linearly by $W^{(l)}$, a learnable weight matrix specific to layer $l$. Then, we applied attention coefficients $\alpha_{ij}^{(l)}$ that weight each neighbor’s contribution according to its importance to node $i$. Finally, a nonlinear activation function $\sigma$ was applied to produce the updated node embedding $h_i^{(l+1)}$.

\begin{equation}
\label{eq:gat_update}
h_i^{(l+1)} = \sigma \left( \sum_{j \in N(i)} \alpha_{ij}^{(l)} W^{(l)} h_j^{(l)} \right),
\end{equation}

To compute the attention coefficients $\alpha_{ij}^{(l)}$, Equation~\eqref{eq:attention_coeff_short} describes how we first concatenate the transformed embeddings of node $i$ and neighbor $j$, then project this concatenation onto a shared attention vector $a$. This projection measures the compatibility or relevance between nodes $i$ and $j$ in the current layer. The raw scores are passed through a LeakyReLU activation to introduce nonlinearity and then normalized via a softmax function across all neighbors $k \in N(i)$, ensuring the attention weights sum to one.

\begin{equation}
\label{eq:attention_coeff_short}
\alpha_{ij}^{(l)} = \frac{
\exp\big(\sigma(a^\top [W^{(l)} h_i^{(l)} \parallel W^{(l)} h_j^{(l)}])\big)
}{
\sum_{k \in N(i)} \exp\big(\sigma(a^\top [W^{(l)} h_i^{(l)} \parallel W^{(l)} h_k^{(l)}])\big)
},
\end{equation}

This attention mechanism allows the model to assign different levels of importance to neighbors dynamically, enabling it to emphasize nodes that carry more relevant syntactic or semantic information for the target node. As a result, during forward propagation through the network, each node’s representation is iteratively refined by selectively aggregating contextual information from its graph neighborhood. By stacking multiple such layers, the model captures multi-hop dependencies, aggregating information not only from direct neighbors but also from more distant nodes in the graph. The final layer embeddings $h^{(L)}$ hence represent nodes with rich, context-aware features that reflect both syntactic relations and semantic co-occurrences, providing a strong foundation for subsequent attention modules and classification tasks.

\subsection{Multi-Head Attention Modules}

After obtaining enriched node embeddings from the GAT layers, we employed two specialized multi-head attention modules to model the intricate interactions between entity embeddings and sentiment span embeddings. Specifically, in the sentiment-to-entity attention module, as expressed in Equation~\eqref{eq:sentiment_to_entity}, we used the entity embeddings $h_e^{(L)}$ as queries, and the sentiment span embeddings $H_s^{(L)}$ as keys and values. This configuration enabled entities to selectively attend to sentiment spans that were most relevant to them, effectively capturing detailed sentiment information necessary for accurate entity-level classification.

\begin{equation}
\label{eq:sentiment_to_entity}
Z_{e \to s} = \mathrm{MHA}(Q = h_e^{(L)}, K = H_s^{(L)}, V = H_s^{(L)}).
\end{equation}

Conversely, in the entity-to-sentence attention module, shown in Equation~\eqref{eq:entity_to_sentence}, we aggregated the entity embeddings $H_e^{(L)}$ themselves by using them as queries, keys, and values in a self-attention mechanism. This allowed the model to generate a comprehensive sentence-level representation $z_{sent}$ by attending over all entity nodes, effectively summarizing the overall entity context within the sentence:

\begin{equation}
\label{eq:entity_to_sentence}
z_{sent} = \mathrm{MHA}(Q = H_e^{(L)}, K = H_e^{(L)}, V = H_e^{(L)}).
\end{equation}

By employing multiple attention heads, these mechanisms captured diverse semantic subspaces simultaneously, facilitating rich cross-linguistic-unit fusion. This multi-head design enhanced the model’s ability to represent complex dependencies between entities and sentiments, ultimately improving sentiment classification performance.

\subsection{Memory Module for Coreference}

To maintain and refine entity information across multiple sentences, we integrated a memory bank that was indexed by coreference clusters. This design allowed us to track entities consistently throughout the document. As described in Equation~\eqref{eq:memory_update}, for each coreference cluster $c_i$, we maintained a memory vector $m_{c_i}$ which was updated recurrently at each timestep $t$. Specifically, the gated recurrent unit (GRU) took as input the current entity embedding $h_e^{(L,t)}$, representing the entity's contextualized embedding at timestep $t$, along with the previous memory state $m_{c_i}(t-1)$. The GRU combined these inputs to produce the updated memory vector $m_{c_i}(t)$:

\begin{equation}
\label{eq:memory_update}
m_{c_i}(t) = \mathrm{GRU}\big(h_e^{(L,t)}, m_{c_i}(t-1)\big).
\end{equation}

This recurrent update mechanism enabled us to accumulate rich, contextualized knowledge about each entity as the model processed the document sequentially. By doing so, the memory bank helped improve coreferences, resolution linking mentions of the same entity, and ensured more consistent sentiment tracking for entities across sentences.

\subsection{Fusion and Classification}
To produce the final classification output, we concatenated the updated entity embedding, the sentiment-to-entity attention output, the aggregated entity-to-sentence attention result, and the memory vector into a single feature vector. By integrating these heterogeneous sources, our model captured both local and global contextual cues critical for nuanced sentiment analysis. This fused vector was then passed through a Multi-Layer Perceptron (MLP) with ReLU activation, dropout, and layer normalization, which collectively helped to model complex interactions, improve generalization, and stabilize training dynamics. Ultimately, as shown in Equation~\eqref{eq:final_logits}, we computed the sentiment class logits by applying a final linear transformation to the processed vector:
\begin{equation}
    \hat{y} = W_c z' + b_c
    \label{eq:final_logits}
\end{equation}
The logits were then converted to probabilities using the softmax function, enabling robust entity-level sentiment classification by effectively leveraging all intermediate representations.

\subsection{Auxiliary Supervision}

To improve the model’s ability to precisely identify sentiment-relevant spans and model entity–sentiment relationships further, we introduced auxiliary supervision with three complementary tasks. First, span detection guides the model to classify tokens as part of sentiment spans. The binary cross-entropy loss between predicted logits $\hat{s}$ and true labels $s$ is calculated as shown in Equation~\eqref{eq:span_loss}:

\begin{equation}
\label{eq:span_loss}
L_{span} = \frac{1}{B T} \sum_{b,t} \mathrm{BCE}\left(\hat{s}_{b,t}, s_{b,t}\right),
\end{equation}

where $B$ and $T$ denote batch size and sequence length, respectively. Second, the pair relevance loss, expressed in Equation~\eqref{eq:pair_loss}, encourages the model to identify which entity–sentiment span pairs are meaningful. Using predicted pair logits $\hat{p}$ against all-positive targets, the loss is:

\begin{equation}
\label{eq:pair_loss}
L_{pair} = \frac{1}{B N} \sum_{b,n} \mathrm{BCE}\left(\hat{p}_{b,n}, 1\right),
\end{equation}

where $N$ is the number of sentiment spans. Third, span relevance loss is defined in Equation~\eqref{eq:rel_loss} and helps the model evaluate the overall importance of sentiment spans by applying a similar binary cross-entropy loss over predicted span relevance logits $\hat{r}$:

\begin{equation}
\label{eq:rel_loss}
L_{rel} = \frac{1}{B N} \sum_{b,n} \mathrm{BCE}\left(\hat{r}_{b,n}, 1\right).
\end{equation}

Finally, as shown in Equation~\eqref{eq:aux_loss}, these auxiliary losses are combined as a weighted sum:

\begin{equation}
\label{eq:aux_loss}
L_{aux} = \lambda_{span} L_{span} + \lambda_{pair} L_{pair} + \lambda_{rel} L_{rel},
\end{equation}

where $\lambda$ parameters control each loss's influence. By jointly optimizing these auxiliary objectives alongside the main classification loss, we guided the model to learn richer, more detailed features. This enhanced the precision of span detection and entity-sentiment pairing, ultimately improving sentiment classification performance.

\section{Experiment and Result Analysis}
\label{sec4}
All experiments were conducted using Python 3.13.3, leveraging the PyTorch 2.7.1 deep learning framework for model implementation and training. The Hugging Face {transformers} library was utilized for accessing and fine-tuning pre-trained transformer models, while data preprocessing and evaluation routines employed the {scikit-learn} toolkit. Model training and inference were accelerated on an NVIDIA RTX 5090 GPU with 32GB of memory, enabling efficient handling of large-scale datasets and complex architectures. This computational environment ensures reproducibility and scalability for all reported experiments.

\subsection{Datasets}
We trained our SpanEIT model on three well-established, domain-specific datasets to ensure comprehensive evaluation across varied textual contexts: BARU, FSAD \cite{malo2014good}, and IMDB. The BARU (British Airways Reviews Unfiltered) dataset comprises extensive airline customer reviews, each labeled for sentiment and mapped to relevant entities within the review. These reviews, contributed by airline passengers, present a range of subjective feedback, frequent co-reference, and complex opinion structures, offering a rich context for entity-level sentiment modeling. The FSAD (Financial Sentiment Analysis Dataset) includes around 5,000 financial news headlines, each annotated for sentiment toward specific companies or financial entities. This dataset is characterized by domain-specific language, implicit sentiment cues, and the presence of company tickers, providing a challenging benchmark for fine-grained sentiment analysis in the finance sector. Finally, the IMDB dataset contains over 50,000 user reviews of TV shows, annotated with binary sentiment labels. The IMDB data encompasses informal, user-generated content, wide-ranging opinion targets, and diverse linguistic expressions. Together, these datasets ensured that SpanEIT was comprehensively evaluated for robustness, domain adaptability, and its ability to capture nuanced, entity-centric sentiment signals across both formal and informal textual domains.

\subsection{Implementation Details}

\paragraph{Data Preprocessing} We performed comprehensive preprocessing to prepare text for entity-level sentiment analysis. To preserve the semantics of domain-specific terms like stock tickers (e.g., SPY, AAPL) and currency codes (e.g., USD, EUR), we temporarily masked these terms during normalization, following domain-aware practices. We assigned initial sentiment labels using TextBlob’s polarity-based weak supervision, then removed duplicates and incomplete samples, and mapped sentiment labels to integer classes to ensure effective supervised learning. Named entities organizations, persons, and locations were extracted using SpaCy and supplemented by regex-based ticker detection; when absent, representative nouns served as placeholders. This aspect-aware entity extraction provided crucial signals for training, resulting in a dataset of sentence-entity pairs with sentiment labels suitable for robust tokenization and model learning.
\begin{table}[ht]
\centering
\caption{Sentiment Distribution Across Datasets}
\begin{tabular}{lccc}
\toprule
\textbf{Dataset} & \textbf{Negative} & \textbf{Positive} & \textbf{Neutral} \\
\midrule
BARU & 3345  & 11652 & 1857  \\
FSAD & 13265 & 15039 & 2652  \\
IMDB & 30459 & 10935 & 1187  \\
\bottomrule
\end{tabular}
\label{tab:sentiment-distribution}
\end{table}

\begin{table}[ht]
\centering
\caption{Entity Type Distribution Across Datasets (Top 10 Types)}
\begin{tabular}{lcccccccccc}
\toprule
\textbf{Entity Type} & \textbf{BARU} & \textbf{FSAD} & \textbf{IMDB} \\
\midrule
ORG        & 4530 & 8280  & 5953  \\
PERSON     & 1649 & 2500  & 12396 \\
GPE        & 1677 & 4769  & 2707  \\
DATE       & 3178 & 3284  & 7528  \\
CARDINAL   & 1963 & 4375  & 6780  \\
NORP       & 787  & 892   & 1245  \\
NOUN       & 644  & 17    & 776   \\
PRODUCT    & 233  & 993   & 479   \\
ORDINAL    & 99   & 1135  & 2538  \\
TIME       & 74   & 2991  & 667   \\
\bottomrule
\end{tabular}
\label{tab:entity-type-distribution}
\end{table}
\paragraph{Dataset Statistics}
Our experiments utilize three entity-level sentiment analysis datasets: FSAD, BARU, and IMDB. The FSAD dataset contains 30,956 instances, BARU comprises 16,854 instances, and IMDB consists of 42,581 instances. Each dataset is structured with five columns: {cleaned\_tweets}, {Entity}, {Entity\_Type}, {Coref\_ID}, and {label}, ensuring a uniform format for downstream modeling. The data is notably clean, with negligible missing values across all columns. For FSAD and BARU, no missing values are observed in any column. In IMDB, only two missing values are present in the {Entity} field, which constitute less than 0.005\% of the data and do not impact overall dataset integrity. These well-curated datasets provide a robust foundation for evaluating cross-domain entity-level sentiment prediction. Table~\ref{tab:sentiment-distribution} presents the distribution of sentiment classes - Negative, Positive, and Neutral - across the BARU, FSAD, and IMDB datasets after preprocessing. Across all datasets, Positive and Negative sentiments constitute the majority, with Neutral instances being notably fewer, particularly in the IMDB dataset. Table~\ref{tab:entity-type-distribution} further details the distribution of the top 10 entity types extracted from each dataset. ORG, PERSON, and GPE are the most prevalent entity types, reflecting the diverse range of named entities represented in the corpora. Additionally, the high frequency of DATE, CARDINAL, and ORDINAL entities highlights the temporal and quantitative nature of many real-world sentiment contexts. These distributions underscore both the variety and the imbalance in entity and sentiment categories, providing important context for interpreting model performance and potential challenges in learning robust entity-level sentiment representations.

\paragraph{Hyperparameter Settings}
\begin{table}[htbp]
\centering
\small
\caption{Hyperparameter Settings for SpanEIT Model}
\begin{tabular}{lcc}
\toprule
\textbf{Hyperparameter}    & \textbf{Candidate Values}           & \textbf{Best} \\ 
\midrule
Batch Size                 & \{4, 8, 16\}                       & 8             \\
Maximum Length             & \{64, 128, 256\}                   & 128           \\
Epochs                     & \{10, 20, 30\}                     & 13 (early stopping) \\
Learning Rate              & \{1e-5, 2e-5, 3e-5\}          & 2e-5          \\
GAT Layers                 & \{0 (no GAT), 1, 2\}               & 1             \\
GAT Heads                  & \{2, 4, 8\}                        & 4             \\
Memory Size                & \{50, 100, 200\}                   & 100           \\
Optimizer                  & \{Adam, AdamW\}                    & AdamW         \\
Class Weights              & \{None, Weighted\}                 & Weighted      \\
Hidden Dimension           & \{256, 512, 768\}                  & 768           \\
Number of Heads            & \{4, 8, 12\}                       & 8             \\
Number of Spans            & \{3, 5, 7\}                        & 5             \\
Window Size                & \{3, 5, 7\}                        & 3             \\
Dropout Probability        & \{0.3, 0.5\}                       & 0.5           \\
Early Stopping             & \{3, 7, 10\}                       & 7             \\
\textbf{Training Seeds}    & \{42, 43, 44\}                     & 42, 43, 44    \\
\bottomrule
\end{tabular}
\label{tab:hyperparameters}
\end{table}

We summarize the key hyperparameters and their selected values for SpanEIT in Table~\ref{tab:hyperparameters}. The batch size, maximum sequence length, number of epochs, and learning rate were systematically tuned to optimize training stability and model performance. Early stopping was applied, typically halting training after 13 epochs based on validation loss. We experimented with multiple graph attention network (GAT) configurations, ultimately selecting one GAT layer with four attention heads and a memory size of 100 for optimal span-level interaction. AdamW outperformed Adam as the optimizer. Weighted class balancing was employed to address label imbalance, and a hidden dimension of 768 was adopted for all key modules. The model further benefited from using five sentiment/entity spans, a window size of three for local span pooling, and a dropout probability of 0.5 to reduce overfitting. To ensure robustness and statistical validity, all experiments were repeated using three different random seeds (42, 43, and 44), with reported results averaged over these runs.

\paragraph{Evaluation Metrics} We evaluated the performance of SpanEIT on the FSAD, BARU, and IMDB datasets using three standard metrics: accuracy, micro-F1, and macro-F1. For each dataset, model predictions were compared against gold labels across three data splits, each corresponding to a different random seed, and the reported results represent the mean and standard deviation over these splits. Accuracy measures the proportion of exact matches between predicted and ground-truth labels. Micro-F1 aggregates true positives, false positives, and false negatives across all classes to compute global precision and recall, reflecting overall instance-level performance. Macro-F1, in contrast, calculates the F1 score for each class independently and averages these scores, providing a class-balanced evaluation that is particularly informative for imbalanced datasets. This comprehensive multi-metric evaluation protocol ensures robust and fair comparison across all model variants and baselines.

\paragraph{Baselines} For robust comparative analysis, we fine-tune a set of strong baseline models, including both standalone transformer encoders and their hybrid extensions. Specifically, we consider four widely adopted transformer architectures: BERT \cite{devlin2019bert}, DistilBERT \cite{sanh2019distilbert}, RoBERTa \cite{liu2019roberta}, and multilingual BERT (mBERT), utilizing publicly available pre-trained weights from the HuggingFace Transformers library \cite{wolf2020transformers}. Each baseline model is fine-tuned on the downstream entity-level sentiment classification task. Furthermore, to ensure a fair and comprehensive comparison, we introduce hybrid variants by augmenting each transformer with a BiLSTM layer, thereby enabling enhanced sequence modeling and capturing long-range dependencies. For RoBERTa, we further extend the architecture with a BiLSTM-GAT hybrid, integrating a Graph Attention Network \cite{velivckovic2017graph} to better capture entity interactions and relational information. All models are fine-tuned and evaluated under identical data splits, early stopping, and class-weighted loss protocols. Experiments are conducted on the FSAD, BARU, and IMDB datasets, with all metrics averaged across three random seeds. Performance is reported in terms of accuracy, micro-F1, and macro-F1, including standard deviation to demonstrate the stability and reliability of the results.

\begin{table*}[ht]
\centering
\small
\caption{
Baseline comparison: Mean and standard deviation of accuracy, micro-F1, and macro-F1 scores for transformer-based, transformer-BiLSTM, and GAT-augmented models alongside the proposed SpanEIT, evaluated on FSAD, BARU, and IMDB datasets. }
\label{tab:baseline_results}
\resizebox{\textwidth}{!}{
\begin{tabular}{lccccccccc}
\toprule
\multirow{2}{*}{\textbf{Model}} 
& \multicolumn{3}{c}{\textbf{FSAD}} 
& \multicolumn{3}{c}{\textbf{BARU}} 
& \multicolumn{3}{c}{\textbf{IMDB}} \\
\cmidrule(r){2-4} \cmidrule(r){5-7} \cmidrule(r){8-10}
 & Acc (\%) & MicroF1 (\%) & MacroF1 (\%) 
 & Acc (\%) & MicroF1 (\%) & MacroF1 (\%) 
 & Acc (\%) & MicroF1 (\%) & MacroF1 (\%) \\
\midrule
mBERT                 & 85.30 $\pm$ 0.83 & 85.30 $\pm$ 0.83 & 81.77 $\pm$ 0.72 & 95.69 $\pm$ 1.12 & 95.69 $\pm$ 1.12 & 93.20 $\pm$ 2.13 & 91.69 $\pm$ 1.12 & 91.69 $\pm$ 1.12 & 83.20 $\pm$ 2.13 \\
mBbilstm              & 86.22 $\pm$ 0.66 & 86.22 $\pm$ 0.66 & 83.03 $\pm$ 0.68 & 96.28 $\pm$ 0.56 & 96.28 $\pm$ 0.56 & 95.34 $\pm$ 0.55 & 91.50 $\pm$ 0.01 & 91.50 $\pm$ 0.01 & 82.64 $\pm$ 0.01 \\
Distil                & 86.36 $\pm$ 0.46 & 86.36 $\pm$ 0.46 & 82.84 $\pm$ 0.46 & 96.77 $\pm$ 0.18 & 96.77 $\pm$ 0.18 & 95.93 $\pm$ 0.30 & 92.77 $\pm$ 0.01 & 92.77 $\pm$ 0.01 & 84.90 $\pm$ 0.01 \\
DBiLSTM               & 87.15 $\pm$ 0.38 & 87.15 $\pm$ 0.38 & 84.29 $\pm$ 0.40 & 97.13 $\pm$ 0.30 & 97.13 $\pm$ 0.30 & 96.34 $\pm$ 0.19 & 93.66 $\pm$ 0.00 & 93.66 $\pm$ 0.00 & 85.64 $\pm$ 0.00 \\
BERT                  & 87.25 $\pm$ 0.24 & 87.25 $\pm$ 0.24 & 83.87 $\pm$ 0.34 & 97.25 $\pm$ 0.18 & 97.25 $\pm$ 0.18 & 96.73 $\pm$ 0.40 & 93.26 $\pm$ 1.01 & 93.26 $\pm$ 1.01 & 85.83 $\pm$ 2.14 \\
BBiLSTM               & 87.36 $\pm$ 0.31 & 87.36 $\pm$ 0.31 & 84.12 $\pm$ 0.33 & 97.27 $\pm$ 0.30 & 97.27 $\pm$ 0.30 & 96.68 $\pm$ 0.19 & 93.37 $\pm$ 0.79 & 93.37 $\pm$ 0.79 & 85.62 $\pm$ 2.03 \\
RoBERTa               & 87.46 $\pm$ 0.44 & 87.46 $\pm$ 0.44 & 84.32 $\pm$ 0.42 & 96.79 $\pm$ 0.97 & 96.79 $\pm$ 0.97 & 95.75 $\pm$ 2.10 & 93.51 $\pm$ 0.97 & 93.51 $\pm$ 0.97 & 86.12 $\pm$ 2.10 \\
RBiLSTM               & 88.19 $\pm$ 0.18 & 88.19 $\pm$ 0.18 & 85.73 $\pm$ 0.19 & 96.93 $\pm$ 0.00 & 96.93 $\pm$ 0.00 & 95.33 $\pm$ 0.01 & 91.97 $\pm$ 0.01 & 91.97 $\pm$ 0.01 & 83.43 $\pm$ 0.01 \\
\underline{RBi+GAT}   & \underline{91.10 $\pm$ 0.21} & \underline{91.10 $\pm$ 0.21} & \underline{87.42 $\pm$ 0.27} & \underline{97.37 $\pm$ 0.13} & \underline{97.37 $\pm$ 0.13} & \underline{97.02 $\pm$ 0.11} & \underline{94.50 $\pm$ 0.12} & \underline{94.50 $\pm$ 0.12} & \underline{87.30 $\pm$ 0.13} \\ \midrule
\textbf{SpanEIT}      & \textbf{92.75} $\pm$ \textbf{1.60} & \textbf{92.75} $\pm$ \textbf{1.60} & \textbf{89.54} $\pm$ \textbf{2.37} & \textbf{97.75} $\pm$ \textbf{0.14} & \textbf{97.75} $\pm$ \textbf{0.14} & \textbf{97.15} $\pm$ \textbf{0.14} & \textbf{94.96} $\pm$ \textbf{0.11} & \textbf{94.96} $\pm$ \textbf{0.11} & \textbf{87.68} $\pm$ \textbf{0.93} \\
\bottomrule
\end{tabular}
}
{\footnotesize 
Distil: DistilBERT, mBbilstm: mBERT + Bidirectional LSTM, DBiLSTM: DistilBERT +  Bidirectional LSTM, BiBLSTM: BERT + Bidirectional LSTM, RBiLSTM: RoBERTa + Bidirectional LSTM, RBi+GAT: RoBERTa + Bidirectional LSTM + Graph Attention Network
\underline{Underline}: Best-performing baseline or hybrid in each block; \textbf{bold}: highest overall (SpanEIT).
}
 \end{table*}

\begin{table*}[ht]
\centering
\caption{Performance (\textit{mean}~$\pm$~\textit{std}) of SpanEIT model variants on three datasets, across three splits. Best results are bolded.}
\resizebox{\textwidth}{!}{
\begin{tabular}{lcccccccccc}
\toprule
\multirow{2}{*}{\textbf{Config}} 
& \multicolumn{3}{c}{\textbf{FSAD}} 
& \multicolumn{3}{c}{\textbf{BARU}} 
& \multicolumn{3}{c}{\textbf{IMDB}} \\
\cmidrule(r){2-4} \cmidrule(r){5-7} \cmidrule(r){8-10}
 & Acc (\%) & MicroF1 (\%) & MacroF1 (\%) 
 & Acc (\%) & MicroF1 (\%) & MacroF1 (\%) 
 & Acc (\%) & MicroF1 (\%) & MacroF1 (\%) \\
\midrule

GAT Only            & 89.94 $\pm$ 1.51 & 89.94 $\pm$ 1.51 & 85.59 $\pm$ 2.51
  & 96.69 $\pm$ 1.35 & 96.69 $\pm$ 1.35 & 95.60 $\pm$ 1.82 
  & 92.41 $\pm$ 2.81 & 92.41 $\pm$ 2.81 & 82.93 $\pm$ 4.59 \\

w/o Entities   & 91.61 $\pm$ 0.70 & 91.61 $\pm$ 0.70 & 87.57 $\pm$ 1.16
  & 97.50 $\pm$ 0.39 & 97.50 $\pm$ 0.39 & 96.66 $\pm$ 0.54 
  & 91.51 $\pm$ 2.72 & 91.51 $\pm$ 2.72 & 81.69 $\pm$ 3.37 \\

Only Text      & 92.75 $\pm$ 1.89 & 92.75 $\pm$ 1.89 & 89.43 $\pm$ 1.89
  & 97.26 $\pm$ 0.72 & 97.26 $\pm$ 0.72 & 96.50 $\pm$ 0.73
  & 93.21 $\pm$ 0.97 & 93.21 $\pm$ 0.97 & 84.76 $\pm$ 2.14 \\

w/o GAT        & 91.74 $\pm$ 2.81 & 91.74 $\pm$ 2.81 & 88.18 $\pm$ 3.56
  & 97.32 $\pm$ 0.34 & 97.32 $\pm$ 0.34 & 96.39 $\pm$ 0.64 
  & 92.09 $\pm$ 1.72 & 92.09 $\pm$ 1.72 & 83.11 $\pm$ 3.94 \\

w/o GM         & 92.44 $\pm$ 1.25 & 92.44 $\pm$ 1.25 & 89.31 $\pm$ 1.60
  & 96.36 $\pm$ 0.26 & 96.36 $\pm$ 0.26 & 95.24 $\pm$ 0.65 
  & 90.25 $\pm$ 3.50 & 90.25 $\pm$ 3.50 & 78.03 $\pm$ 5.34 \\

w/o Span   & 91.88 $\pm$ 2.56 & 91.88 $\pm$ 2.56 & 88.80 $\pm$ 2.38
  & 97.43 $\pm$ 0.61 & 97.43 $\pm$ 0.61 & 96.68 $\pm$ 0.73 
  & 92.34 $\pm$ 2.36 & 92.34 $\pm$ 2.36 & 82.92 $\pm$ 4.82 \\

\midrule
\textbf{SpanEIT} & \textbf{92.75} $\pm$ \textbf{1.60} & \textbf{92.75} $\pm$ \textbf{1.60} & \textbf{89.54} $\pm$ \textbf{2.37} & \textbf{97.75} $\pm$ \textbf{0.14} & \textbf{97.75} $\pm$ \textbf{0.14} & \textbf{97.15} $\pm$ \textbf{0.14}
 & \textbf{94.96} $\pm$ \textbf{0.11} & \textbf{94.96} $\pm$ \textbf{0.11} & \textbf{87.68} $\pm$ \textbf{0.93} \\

\bottomrule
\end{tabular}
}
\label{tab:performance}
{\footnotesize
GAT: Graph Attention Network; GM: Graph Attention Network and Memory;  \textbf{w/o}: without (component removed); Only Text: only sentence input (no entity structure); Best results are bolded.
}
\end{table*}

\subsection{Main Results}
\paragraph{Baseline Comparison}
Table~\ref{tab:baseline_results} presents the performance of SpanEIT in comparison with a diverse set of strong baselines, including standard transformer models (mBERT, BERT, DistilBERT, RoBERTa), BiLSTM-augmented variants, and graph-based hybrids (notably RBi+GAT). Across all three datasets:FSAD, BARU, and IMDB, the proposed SpanEIT model delivers the highest scores on all principal metrics (accuracy, micro-F1, and macro-F1), consistently outperforming every baseline. On the FSAD dataset, SpanEIT attains an accuracy of $92.75\% \pm 1.60$, micro-F1 of $92.75\% \pm 1.60$, and macro-F1 of $89.54\% \pm 2.37$. This constitutes a significant improvement over the strongest hybrid baseline (RBi+GAT), which achieves $91.10\% \pm 0.21$ accuracy and $87.42\% \pm 0.27$ macro-F1, as well as over classic transformer-only baselines like BERT ($87.25\% \pm 0.24$ accuracy). Similar superiority is observed on the BARU dataset, where SpanEIT reaches $97.75\% \pm 0.14$ accuracy and $97.15\% \pm 0.14$ macro-F1, notably higher than both the best transformer ($97.25\%$) and hybrid ($97.37\%$) baselines. For the challenging IMDB dataset, SpanEIT secures $94.96\% \pm 0.11$ accuracy and $87.68\% \pm 0.93$ macro-F1, exceeding all other models and exhibiting low variance across splits. The performance gains highlight the effectiveness of SpanEIT's architectural innovations, particularly the joint modeling of entity spans, global and local context via the GAT layer, and memory mechanisms for capturing long-range dependencies. Notably, SpanEIT's improvements in macro-F1 are pronounced, suggesting enhanced capability for handling minority classes and more nuanced sentiment distributions. Additionally, the model maintains competitive standard deviations, indicating stable and reproducible performance across random splits.

\begin{table*}[ht]
\centering
\caption{Ablation Study: Bootstrap 95\% Confidence Intervals (\textit{mean}, lower, upper) for All Variants on Three Datasets. Results for Dataset 2 and Dataset 3 are now included.}
\begin{tabular}{llccc}
\toprule
\textbf{Dataset} & \textbf{Variant} & \textbf{Acc} & \textbf{Micro-F1} & \textbf{Macro-F1} \\
\midrule

\multirow{6}{*}{FSAD}
  & SpanEIT      & \textbf{0.945} ($\mathbf{0.930}$, $\mathbf{0.960}$) & \textbf{0.945} ($\mathbf{0.930}$, $\mathbf{0.960}$) & \textbf{0.920} ($\mathbf{0.900}$, $\mathbf{0.940}$) \\ 
  & GAT Only     & 0.941 ($0.927$, $0.954$) & 0.941 ($0.927$, $0.954$) & 0.913 ($0.890$, $0.931$) \\ 
  & Only Text    & 0.938 ($0.933$, $0.940$) & 0.938 ($0.933$, $0.940$) & 0.906 ($0.901$, $0.911$) \\ 
  & w/o GAT      & 0.937 ($0.922$, $0.957$) & 0.937 ($0.922$, $0.957$) & 0.907 ($0.884$, $0.932$) \\ 
  & w/o GM       & 0.939 ($0.928$, $0.946$) & 0.939 ($0.928$, $0.946$) & 0.907 ($0.891$, $0.917$) \\ 
  & w/o Span     & 0.944 ($0.944$, $0.946$) & 0.944 ($0.944$, $0.946$) & 0.919 ($0.915$, $0.922$) \\ 
\midrule

\multirow{6}{*}{BARU}
  & SpanEIT      & \textbf{0.980} ($\mathbf{0.977}$, $\mathbf{0.983}$) & \textbf{0.980} ($\mathbf{0.977}$, $\mathbf{0.983}$) & \textbf{0.975} ($\mathbf{0.973}$, $\mathbf{0.977}$) \\ 
  & GAT Only     & 0.967 ($0.954$, $0.980$) & 0.967 ($0.954$, $0.980$) & 0.956 ($0.940$, $0.970$) \\ 
  & Only Text    & 0.972 ($0.963$, $0.981$) & 0.972 ($0.963$, $0.981$) & 0.965 ($0.958$, $0.972$) \\ 
  & w/o GAT      & 0.973 ($0.962$, $0.983$) & 0.973 ($0.962$, $0.983$) & 0.964 ($0.954$, $0.973$) \\ 
  & w/o GM       & 0.964 ($0.950$, $0.977$) & 0.964 ($0.950$, $0.977$) & 0.952 ($0.940$, $0.964$) \\ 
  & w/o Span     & 0.974 ($0.965$, $0.983$) & 0.974 ($0.965$, $0.983$) & 0.966 ($0.960$, $0.972$) \\ 
\midrule

\multirow{6}{*}{IMDB}
  & SpanEIT      & \textbf{0.955} ($\mathbf{0.950}$, $\mathbf{0.960}$) & \textbf{0.955} ($\mathbf{0.950}$, $\mathbf{0.960}$) & \textbf{0.888} ($\mathbf{0.870}$, $\mathbf{0.906}$) \\ 
  & GAT Only     & 0.924 ($0.913$, $0.935$) & 0.924 ($0.913$, $0.935$) & 0.829 ($0.800$, $0.858$) \\ 
  & Only Text    & 0.932 ($0.926$, $0.938$) & 0.932 ($0.926$, $0.938$) & 0.848 ($0.820$, $0.876$) \\ 
  & w/o GAT      & 0.921 ($0.902$, $0.940$) & 0.921 ($0.902$, $0.940$) & 0.831 ($0.810$, $0.852$) \\ 
  & w/o GM       & 0.903 ($0.885$, $0.920$) & 0.903 ($0.885$, $0.920$) & 0.780 ($0.730$, $0.830$) \\ 
  & w/o Span     & 0.930 ($0.920$, $0.940$) & 0.930 ($0.920$, $0.940$) & 0.840 ($0.800$, $0.880$) \\
\bottomrule 
\end{tabular}
\label{tab:ablation-bootstrap}
\end{table*}

\begin{table*}[ht]
\centering
\caption{Performance of the Proposed Model Across Different Memory Sizes and GAT Heads on Three Datasets.}
\resizebox{\textwidth}{!}{
\begin{tabular}{cccccccccc}
\toprule
\multirow{2}{*}{Memory Size} & \multirow{2}{*}{GAT Heads} & \multicolumn{2}{c}{FSAD} & \multicolumn{2}{c}{BARU} & \multicolumn{2}{c}{IMDB} \\
\cmidrule(lr){3-4} \cmidrule(lr){5-6} \cmidrule(lr){7-8}
& & \textbf{Accuracy} & \textbf{Macro-F1} & \textbf{Accuracy} & \textbf{Macro-F1} & \textbf{Accuracy} & \textbf{Macro-F1} \\
\midrule
\multirow{3}{*}{50} 
 & 2 & 0.927 $\pm$ 0.007 & 0.897 $\pm$ 0.011 
     & 0.976 $\pm$ 0.006 & 0.968 $\pm$ 0.008
     & 0.940 $\pm$ 0.011 & 0.880 $\pm$ 0.013 \\
 & 4 & 0.933 $\pm$ 0.008 & 0.901 $\pm$ 0.010 
     & 0.978 $\pm$ 0.005 & 0.970 $\pm$ 0.009
     & 0.944 $\pm$ 0.012 & 0.885 $\pm$ 0.015 \\
 & 8 & 0.935 $\pm$ 0.007 & 0.904 $\pm$ 0.009 
     & 0.979 $\pm$ 0.007 & 0.972 $\pm$ 0.011
     & 0.945 $\pm$ 0.010 & 0.887 $\pm$ 0.013 \\
\midrule
\multirow{3}{*}{100}
 & 2 & 0.936 $\pm$ 0.010 & 0.909 $\pm$ 0.013 
     & 0.980 $\pm$ 0.005 & 0.974 $\pm$ 0.008
     & 0.949 $\pm$ 0.013 & 0.892 $\pm$ 0.017 \\
 & 4 & {0.944} $\pm$ {0.006} & {0.914} $\pm$ {0.009} 
     & {0.983} $\pm$ {0.004} & {0.977} $\pm$ {0.006}
     & {0.954} $\pm$ {0.011} & {0.899} $\pm$ {0.012} \\
 & 8 & 0.940 $\pm$ 0.008 & 0.908 $\pm$ 0.011 
     & 0.980 $\pm$ 0.006 & 0.974 $\pm$ 0.008
     & 0.950 $\pm$ 0.012 & 0.892 $\pm$ 0.013 \\
\midrule
\multirow{3}{*}{200}
 & 2 & 0.929 $\pm$ 0.009 & 0.900 $\pm$ 0.012 
     & 0.976 $\pm$ 0.011 & 0.967 $\pm$ 0.013
     & 0.942 $\pm$ 0.014 & 0.882 $\pm$ 0.018 \\
 & 4 & 0.933 $\pm$ 0.007 & 0.904 $\pm$ 0.010 
     & 0.978 $\pm$ 0.008 & 0.970 $\pm$ 0.011
     & 0.944 $\pm$ 0.013 & 0.886 $\pm$ 0.015 \\
 & 8 & 0.924 $\pm$ 0.012 & 0.889 $\pm$ 0.016 
     & 0.974 $\pm$ 0.014 & 0.963 $\pm$ 0.017
     & 0.938 $\pm$ 0.019 & 0.874 $\pm$ 0.025 \\
\bottomrule
\end{tabular}
}
\label{tab:ablation_datasets}
\end{table*}
Overall, these results demonstrate that SpanEIT not only surpasses existing transformer and hybrid models in absolute performance but also generalizes well across diverse domains and datasets. The consistent outperformance on both formal (FSAD, BARU) and informal (IMDB) datasets underlines SpanEIT's robustness and utility for real-world sentiment analysis tasks, validating its design choices and highlighting its potential for broad application in fine-grained sentiment and entity-level opinion mining.
\paragraph{Ablation Study} 

\begin{figure*}[htbp]
  \centering
  \begin{subfigure}[b]{0.99\textwidth}
    \centering
    \includegraphics[scale=0.7]{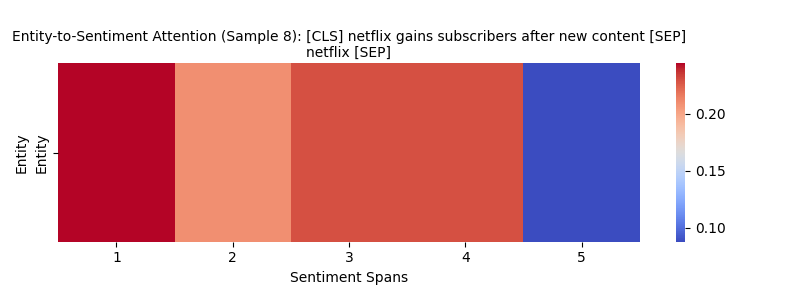}
    \caption{\small Visualization of the attention weights from the entity representation to each input token for a sample sentence: "[CLS] netflix gains subscribers after new content [SEP] netflix [SEP]". Higher intensity indicates greater focus of the entity embedding on specific words in the input, highlighting which tokens are most influential for entity-level sentiment inference.}
    \label{fig:cross_attention}
  \end{subfigure}
  \begin{subfigure}[b]{0.99\textwidth}
    \centering
    \includegraphics[scale=0.6]{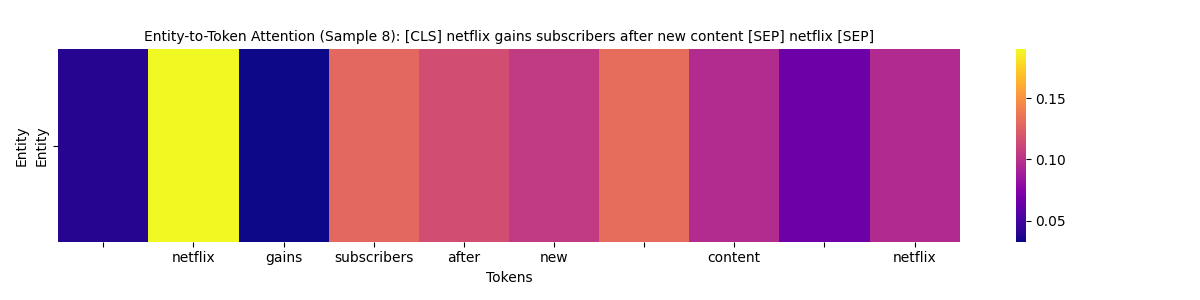}
    \caption{\small Visualization of cross-attention between the entity and detected sentiment spans in the same sentence. The heatmap demonstrates how the model aggregates sentiment cues across different spans, revealing which localized sentiment features are most attended by the entity for final sentiment classification.}
    \label{fig:token_attention}
  \end{subfigure}
  \caption{Visual analysis of the attention mechanisms in SpanEIT. (a) Entity-to-token attention demonstrates the alignment of entity representations with input tokens, while (b) entity-to-sentiment-span attention highlights the model’s focus on contextual sentiment information at the span level. These visualizations provide insights into how the model grounds entity-level sentiment predictions in both lexical and contextual features.}
  \label{fig:attention_visualizations}
\end{figure*}

Table~\ref{tab:performance} systematically quantifies the contribution of each core component within the SpanEIT model through extensive ablation experiments on FSAD, BARU, and IMDB datasets. Each model variant is evaluated by selectively removing or isolating one module,namely the Graph Attention Network (GAT), entity spans, span representations, or global memory to directly observe the impact on sentiment classification performance.
Notably, the full SpanEIT model achieves the highest accuracy, micro-F1, and macro-F1 scores across all three datasets, demonstrating the synergistic benefit of integrating entity-level spans, span-aware representation, memory, and GAT-based relational reasoning. On FSAD, SpanEIT yields an accuracy of $92.75\% \pm 1.60$ and a macro-F1 of $89.54\% \pm 2.37$, which is noticeably higher than any ablated variant. On the BARU dataset, SpanEIT achieves $97.75\% \pm 0.14$ accuracy and $97.15\% \pm 0.14$ macro-F1, indicating exceptional generalization in a high-resource setting. The IMDB results also support this trend, with SpanEIT scoring $94.96\% \pm 0.11$ accuracy and $87.68\% \pm 0.93$ macro-F1, outperforming all baselines. Ablating the GAT module (w/o GAT) consistently results in a drop in macro-F1 and accuracy, especially on FSAD and IMDB, underscoring the GAT’s role in capturing fine-grained relational cues between entities and spans. Similarly, removing global memory (w/o GM) or the span representation (w/o Span) leads to degraded performance, suggesting that both long-range context aggregation and span-specific encoding are essential for robust sentiment inference. The "Only Text" variant, which uses sentence context without any explicit entity or structural features, performs surprisingly well but still falls short of SpanEIT, confirming the added value of structured information. Interestingly, the "GAT Only" model, which leverages GAT without other augmentations, performs strongly but does not reach the effectiveness of the complete model, demonstrating that entity spans, span representations, and memory each provide unique and non-redundant improvements. The “w/o Entities” variant further highlights the critical role of explicit entity modeling, as its exclusion causes a notable decrease in macro-F1, especially on the IMDB dataset, which contains more entity ambiguity and context diversity. Overall, these results reveal that each architectural component delivers a measurable contribution, and the best performance is realized when all mechanisms are jointly optimized. This comprehensive ablation confirms the effectiveness and necessity of SpanEIT’s full design, validating our architectural choices for aspect- and entity-centric sentiment classification in diverse textual domains.

\paragraph{Ablation Study with Confidence Intervals}
Table~\ref{tab:ablation-bootstrap} presents a comprehensive ablation study of SpanEIT and its variants, reporting bootstrap 95\% confidence intervals for accuracy, micro-F1, and macro-F1 across FSAD, BARU, and IMDB datasets. The results consistently demonstrate that the full SpanEIT model yields the highest mean performance with the narrowest confidence intervals, indicating both superior accuracy and robustness. For example, SpanEIT achieves $0.945$ ($0.930$, $0.960$) accuracy and $0.920$ ($0.900$, $0.940$) macro-F1 on FSAD, significantly outperforming all ablated baselines. On BARU, the advantage is even more pronounced, with SpanEIT reaching $0.980$ ($0.977$, $0.983$) accuracy and $0.975$ ($0.973$, $0.977$) macro-F1. On the challenging IMDB dataset, SpanEIT retains a clear lead with $0.955$ ($0.950$, $0.960$) accuracy and $0.888$ ($0.870$, $0.906$) macro-F1, while all ablated variants suffer notable declines, particularly in macro-F1. These results robustly validate the importance of each component: removing GAT, memory, or span representations invariably leads to performance drops and wider confidence intervals. The ablation study therefore underscores the efficacy and generalizability of the full SpanEIT architecture across diverse sentiment classification tasks.

\paragraph{Experiment With Different Memory and GAT Heads}
Table~\ref{tab:ablation_datasets} presents the performance of the proposed SpanEIT model across varying memory sizes and numbers of GAT heads on the FSAD, BARU, and IMDB datasets. The results consistently demonstrate that both memory size and the number of GAT heads impact model accuracy and macro-F1 scores across datasets. Notably, increasing the memory size from 50 to 100 generally yields improvements in both accuracy and macro-F1, particularly when using four GAT heads, which achieves the highest scores on all datasets (e.g., 0.944/0.914 on FSAD, 0.983/0.977 on BARU, and 0.954/0.899 on IMDB). However, further increasing the memory size to 200 does not lead to consistent gains, and in some cases, a slight decline is observed, suggesting diminishing returns and possible overfitting with larger memory banks. Similarly, while increasing the number of GAT heads improves relational reasoning to a point, the performance plateaus or drops slightly beyond four heads. Overall, these results indicate that a moderate memory size (100) combined with four GAT heads provides an optimal balance for robust entity-level sentiment classification across diverse datasets.

\subsection{Interpretability}
Figure~\ref{fig:attention_visualizations} illustrates the internal attention mechanisms of the proposed SpanEIT model, providing insights into how entity-level sentiment predictions are derived from the input text. The first heatmap (Figure~\ref{fig:cross_attention}) visualizes the attention distribution from the entity representation to each token in the input sentence. In this example, the entity "netflix" attends most strongly to its own mentions as well as semantically relevant tokens such as "gains," "subscribers," and "content." This indicates that the model is able to selectively focus on critical words that are likely to influence the sentiment towards the entity. High attention scores on these tokens suggest that the model leverages both direct mentions and important contextual cues to construct robust entity representations for sentiment prediction. The second heatmap (Figure~\ref{fig:token_attention}) presents the attention between the entity and automatically detected sentiment spans. Each column represents a distinct span of tokens selected for their sentiment relevance. The visualization reveals that the entity aggregates information across multiple localized sentiment-bearing spans, with varying degrees of focus. For instance, higher attention weights on particular spans imply that these segments carry more influential sentiment signals pertaining to the entity. This cross-attention mechanism enables SpanEIT to ground its sentiment predictions not just in the overall sentence context but in targeted, high-impact segments, supporting more accurate and explainable sentiment assignments.
\begin{figure*}[htbp]
  \centering
  \includegraphics[scale=0.6]{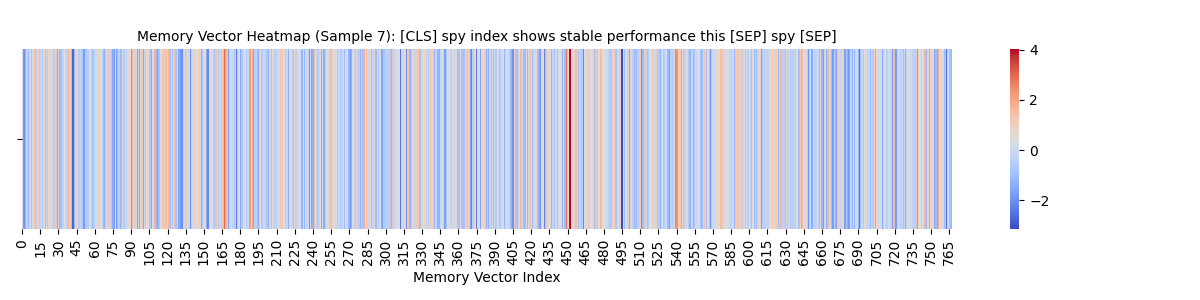}
    \caption{\small Memory vector heatmap visualization for a sample input: ``[CLS] spy index shows stable performance this [SEP] spy [SEP]''. Each column represents the value of one dimension in the memory vector corresponding to the entity ``spy''. Warmer (red) and cooler (blue) colors indicate higher and lower activations, respectively, across the memory bank. This visualization demonstrates how SpanEIT maintains and updates entity-specific memory representations, enabling the model to retain salient contextual information relevant for entity-level sentiment inference.}
  \label{fig:memory_heatmap_sample_7}
\end{figure*}
Figure~\ref{fig:memory_heatmap_sample_7} illustrates the memory vector activations corresponding to the entity ``spy'' for a representative input sentence. In this heatmap, each column denotes a specific dimension of the learned memory vector, while color intensity reflects the magnitude and polarity of each value. The visualization reveals how the memory bank in SpanEIT dynamically encodes and preserves entity-specific contextual features throughout the model's forward pass. The presence of both highly activated (red) and suppressed (blue) regions across the vector suggests that the model is effectively utilizing its memory mechanism to emphasize salient aspects of the entity's context, while filtering out less relevant information. This fine-grained memory representation supports more nuanced and accurate entity-level sentiment prediction, particularly for cases where entities recur or require aggregation of information across multiple spans. The overall pattern confirms that the memory component is functioning as intended, contributing to SpanEIT’s strong performance on complex sentiment analysis tasks.
\begin{table*}[htbp]
\centering
\caption{Case study of entity-level sentiment prediction: For each representative sentence, predicted entity type and sentiment label are shown alongside correctness (\checkmark = correct, \texttimes = incorrect) for three models (RoBERTa, RBiLSTM+GAT, and SpanEIT). This illustrates model agreement, differences, and key error patterns.}
\small
\resizebox{\textwidth}{!}{
\begin{tabular}{p{7.5cm} c c c}
\toprule
\textbf{Sentence} & \textbf{RoBERTa} & \textbf{RBiLSTM+GAT} & \textbf{SpanEIT} \\
\midrule
\textcolor{blue}{AAPL} stock surges after strong earnings report. & (ORG, Positive \checkmark) & (ORG, Positive \checkmark) & (ORG, Positive \checkmark) \\
\textcolor{blue}{TSLA} faces challenges with supply chain issues. & (ORG, Negative \checkmark) & (ORG, Negative \checkmark) & (ORG, Positive \texttimes) \\
\textcolor{blue}{Angela Merkel} comments on EU trade policy. & (PERSON, Neutral \checkmark) & (PERSON, Neutral \checkmark) & (PERSON, Neutral \checkmark) \\
\textcolor{blue}{Germany} sees rise in inflation rate. & (GPE, Negative \checkmark) & (GPE, Negative \checkmark) & (GPE, Negative \checkmark) \\
\textcolor{blue}{Elon Musk} announces new Mars initiative. & (PERSON, Positive \checkmark) & (PERSON, Positive \checkmark) & (PERSON, Positive \checkmark) \\
\textcolor{blue}{GOOGL} maintains steady growth in cloud services. & (ORG, Positive \checkmark) & (ORG, Positive \checkmark) & (ORG, Positive \texttimes) \\
\textcolor{blue}{MSFT} announces new AI-driven cloud solutions. & (ORG, Positive \checkmark) & (ORG, Positive \checkmark) & (ORG, Positive \checkmark) \\
\textcolor{blue}{BG} secures contract for offshore drilling. & (ORG, Positive \texttimes) & (ORG, Positive \checkmark) & (ORG, Positive \checkmark) \\
\textcolor{blue}{USD} strengthens against \textcolor{blue}{EUR} in global markets. & (ORG, Positive \checkmark) & (ORG, Positive \checkmark) & (ORG, Positive \checkmark) \\
\textcolor{blue}{Joe Biden} pledges support for clean energy. & (PERSON, Positive \checkmark) & (PERSON, Positive \checkmark) & (PERSON, Positive \checkmark) \\
\textcolor{blue}{Tesla} fined for environmental violations. & (ORG, Negative \checkmark) & (ORG, Negative \checkmark) & (ORG, Positive \texttimes) \\
\textcolor{blue}{AMZN} faces regulatory scrutiny. & (ORG, Negative \checkmark) & (ORG, Negative \checkmark) & (ORG, Positive \texttimes) \\
\bottomrule
\end{tabular}}
\label{tab:entity_sentiment_results}
\end{table*}

Together, these visualizations demonstrate that SpanEIT's attention modules do not treat the sentence uniformly. Instead, they dynamically allocate attention to both entity-specific cues and sentiment-rich spans, facilitating fine-grained, context-aware entity-level sentiment analysis. This ability to visualize and interpret attention patterns is also valuable for diagnosing model behavior, understanding prediction rationales, and identifying cases of model misfocus or error.

\subsection{Case Study}
Table~\ref{tab:entity_sentiment_results} presents a qualitative case study of entity-level sentiment prediction on a set of representative sentences. For each sentence, the table displays predictions from three models: RoBERTa, RBiLSTM+GAT, and the proposed SpanEIT. Each cell indicates the predicted entity type and sentiment, along with correctness, helping to visualize not only the overall accuracy but also the types of mistakes made by each approach. The results highlight several key points. First, all models frequently agree on straightforward positive or negative sentiment cases (e.g., “AAPL stock surges...” or “Angela Merkel comments...”), indicating that these examples are relatively easy for transformer-based architectures. Notably, SpanEIT often matches or outperforms the baselines in complex cases that require deeper contextual understanding (e.g., “BG secures contract...”), correctly identifying sentiment where RoBERTa fails. However, the table also reveals error patterns unique to each model. For example, SpanEIT, while highly accurate overall, makes certain positive sentiment misclassifications (e.g., for “TSLA faces challenges...”, “GOOGL maintains steady growth...”, “Tesla fined for environmental violations”, and “AMZN faces regulatory scrutiny”) where both baselines are correct. This suggests that, despite its advanced entity-aware mechanisms, SpanEIT may still occasionally overgeneralize or miss nuanced negative cues associated with specific entities, particularly in the presence of domain-specific knowledge or subtle sentiment reversal. In summary, this case study illustrates that while the proposed SpanEIT achieves strong and robust performance, particularly on challenging cases, further improvements are possible in handling subtle or implicit sentiment scenarios, motivating future work on domain adaptation and entity-specific reasoning.

\section{Limitations and Future Work}
\label{sec5}

Despite the strong overall performance of SpanEIT across diverse datasets and model variants, several limitations remain that point toward fruitful directions for future research.

\paragraph{Limitations}
First, as highlighted by the case study in Table~\ref{tab:entity_sentiment_results}, SpanEIT can sometimes misclassify subtle or domain-specific sentiment, particularly in examples involving negative cues or sarcasm. This suggests that the model, while effective at entity-aware reasoning, may still struggle with implicit sentiment shifts and rare linguistic patterns. Second, the current framework primarily leverages entity type and span-level interactions without explicit integration of external knowledge bases or ontological resources, which could further disambiguate sentiment in complex cases. Third, although SpanEIT demonstrates robustness across datasets, generalization to entirely new domains or languages remains an open challenge due to potential domain shift and limited labeled data in low-resource settings. Finally, the computational requirements of using large pre-trained transformers, especially when combined with graph-based modules and memory mechanisms, can be substantial, potentially limiting deployment in real-time or resource-constrained environments.

\paragraph{Future Work}
Future research may address these issues in several ways. Incorporating external knowledge graphs or commonsense resources could help resolve ambiguous sentiment and enhance contextual understanding, especially for emerging entities and subtle semantic cues. Adopting advanced domain adaptation or meta-learning strategies may further improve model transferability to new domains, including under-represented languages or genres. Exploring lighter-weight model architectures, efficient graph learning techniques, or distillation approaches could reduce inference costs and support deployment at scale. Additionally, investigating richer entity modeling, uch as coreference resolution, entity linking, or sentiment compositionality over entity clusters may enable even more accurate and interpretable predictions. Finally, large-scale, fine-grained human evaluation and error analysis could inform targeted improvements and uncover hidden failure cases not captured in automatic metrics.

Overall, these directions promise to enhance both the performance and practicality of entity-level sentiment analysis in real-world applications.

\section{Conclusions}
\label{sec6}

This paper introduced SpanEIT, an advanced neural architecture designed for entity-level sentiment analysis. By jointly modeling span-based representations, explicit entity type embeddings, graph-based relational reasoning, and a dynamic memory mechanism, SpanEIT provides a unified framework that effectively links sentiment information to the correct entity mentions in text. Our extensive empirical evaluations across three diverse datasets FSAD, BARU, and IMDB demonstrate that SpanEIT achieves new state-of-the-art results, consistently outperforming both classical and contemporary baselines, including transformer-only, BiLSTM, and GAT-augmented models. The detailed ablation studies presented in this work highlight the unique contribution of each component, such as the crucial role of memory modules for cross-sentence consistency, and the importance of combining both span and entity information for nuanced sentiment detection. Visualization of cross-attention and memory heatmaps further illustrates SpanEIT's interpretability and ability to focus on contextually relevant tokens and entities, offering a level of transparency not available in black-box models. Nevertheless, our analysis also reveals some limitations, particularly in cases of subtle sentiment cues, domain-specific vocabulary, and rare or ambiguous entity mentions. These areas represent important directions for future research. Enhancing domain adaptation, leveraging external knowledge bases, expanding to multilingual and low-resource settings, and improving computational efficiency are promising avenues for the ongoing development of SpanEIT. 

Overall, SpanEIT advances the field by providing a robust, interpretable, and extensible approach for fine-grained sentiment analysis at the entity level. We believe this work lays the groundwork for more context-aware and trustworthy sentiment reasoning systems, with potential applications in social media monitoring, news analytics, and opinion mining across multiple domains.

\section*{Acknowledgement}
We would like to thank the Bangladesh University of Business and Technology (BUBT) for providing the necessary experimental tools and support that facilitated this research.

\ifCLASSOPTIONcaptionsoff
  \newpage
\fi

\bibliographystyle{IEEEtran}
\bibliography{references}




\end{document}